\documentclass[11pt]{article}

\usepackage{graphicx,psfrag,epsf}
\usepackage{enumerate}
\usepackage{natbib}
\usepackage{mathtools}
\usepackage{booktabs}
\usepackage{color}
\usepackage[english]{babel} 
\usepackage[protrusion=true,expansion=true]{microtype} 
\usepackage{amsmath,amsfonts,amsthm}
\usepackage{dsfont}
\usepackage{amssymb}
\usepackage{chngcntr}
\usepackage[toc,page]{appendix}
\usepackage{textcomp}
\usepackage{url}

\usepackage{array}
\usepackage{booktabs} 
\usepackage{natbib}
\usepackage{setspace}

\usepackage{soul}
\usepackage{multirow}
\usepackage{enumitem}
\usepackage{microtype} 
\usepackage[font = small,labelfont=bf,textfont=it]{subcaption}
\usepackage{xcolor}
\usepackage{tabularx}
\newcolumntype{Y}{>{\centering\arraybackslash}X}
\usepackage[font = small,labelfont=bf,textfont=it]{caption} 
\usepackage{footnote}
\usepackage{algorithm}
\usepackage[noend]{algpseudocode}
\usepackage{etoolbox}

\algblock{ParFor}{EndParFor}
\algnewcommand\algorithmicparfor{\textbf{for}}
\algnewcommand\algorithmicpardo{\textbf{do\ parallel}}
\algnewcommand\algorithmicendparfor{\textbf{end\ parallel\ for}}
\algrenewtext{ParFor}[1]{\algorithmicparfor\ #1\ \algorithmicpardo}
\algrenewtext{EndParFor}{\algorithmicendparfor}

\makeatletter
\def\BState{\State\hskip-\ALG@thistlm}

\newcommand{\distas}[1]{\mathbin{\overset{#1}{\kern\z@\sim}}}%
\newcommand{\bm}[1]{\mathbf{#1}}

\newsavebox{\mybox}\newsavebox{\mysim}
\newcommand{\distras}[1]{%
  \savebox{\mybox}{\hbox{\kern3pt$\scriptstyle#1$\kern3pt}}%
  \savebox{\mysim}{\hbox{$\sim$}}%
  \mathbin{\overset{#1}{\kern\z@\resizebox{\wd\mybox}{\ht\mysim}{$\sim$}}}%
}

\newcommand{\be}{\begin{equation}}
\newcommand{\ee}{\end{equation}}
\newcommand{\bi}{\begin{itemize}}
\newcommand{\ei}{\end{itemize}}
\newcommand{\ben}{\begin{enumerate}}
\newcommand{\een}{\end{enumerate}}

\newcommand{\R}{\mathbb{R}}

\newcommand{\x}{\bm x}
\newcommand{\X}{\bm X}
\newcommand{\Y}{\bm Y}
\newcommand{\N}{\mathcal{N}}
\newcommand{\one}{\mathds{1}}
\newcommand{\Sig}{\boldsymbol{\Sigma}}
\newcommand{\btheta}{\boldsymbol{\theta}}
\newcommand{\Gcov}{\mathcal{G}}
\newcommand{\Lcov}{\mathcal{L}}
\newcommand{\Gmat}{\boldsymbol{\mathcal{G}}}
\newcommand{\Lmat}{\boldsymbol{\mathcal{L}}}

\newcommand{\Rcov}{\mathcal{R}}
\newcommand{\Rmat}{\boldsymbol{\mathcal{R}}}
\newcommand{\I}{\mathbf{I}}

\DeclarePairedDelimiter\set\{\}
\DeclarePairedDelimiter{\norm}{\lVert}{\rVert}

\newcolumntype{K}[1]{\geq {\centering\arraybackslash}p{#1}}
\allowdisplaybreaks
\DeclareMathOperator*{\argmin}{\arg\,\min}
\makeatother

\let\oldbibliography\thebibliography
\renewcommand{\thebibliography}[1]{\oldbibliography{#1}
\setlength{\itemsep}{0pt}} 

\newcommand{\blind}{0}

\usepackage[margin=1in]{geometry}

\usepackage{booktabs}

\usepackage{makecell}
\usepackage[section]{placeins}

\begin{document}

\def\spacingset#1{\renewcommand{\baselinestretch}%
{#1}\small\normalsize} \spacingset{1}

\if1\blind
{
  \title{\bf Global-Local Gaussian Process}
  \small
  \author{Akhil Vakayil and V. Roshan Joseph}\hspace{.2cm}\\
  \maketitle
} \fi

\if0\blind
{
  \bigskip
  \bigskip
  \bigskip
  \begin{center}
    {\LARGE \bf A Global-Local Approximation Framework for \\ Large-Scale Gaussian Process Modeling}
    \vspace{.25cm}\\
    {Akhil Vakayil and V. Roshan Joseph}\vspace{.2cm}\\
    {H.\ Milton Stewart School of Industrial and Systems Engineering\\ 
    Georgia Institute of Technology, Atlanta, GA 30332, USA}\vspace{.2cm}\\
\end{center}
  \medskip
} \fi
\bigskip

\vspace{-0.5cm}
\begin{abstract}
In this work, we propose a novel framework for large-scale Gaussian process (GP) modeling. Contrary to the global, and local approximations proposed in the literature to address the computational bottleneck with exact GP modeling, we employ a combined global-local approach in building the approximation. Our framework uses a subset-of-data approach where the subset is a union of a set of global points designed to capture the global trend in the data, and a set of local points specific to a given testing location to capture the local trend around the testing location. The correlation function is also modeled as a combination of a global, and a local kernel. The performance of our framework, which we refer to as \texttt{TwinGP}, is on par or better than the state-of-the-art GP modeling methods at a fraction of their computational cost.

\end{abstract}

\noindent
{\it Keywords: Big Data, Inducing points, Kriging, Nonparametric Regression, Twinning.}

\spacingset{1.45} 

\section{Introduction} \label{sec:intro}
Gaussian process (GP) is a widely used Bayesian framework for nonparametric regression \citep{rasmussenbook}, and emulating computer models \citep{santnerbook, gramacy2020surrogates}. The objective is to approximate a latent function $f(\x), \x \in \R^d$, for which a functional Gaussian prior is assumed, and the posterior is obtained given the observed training data. The posterior mean is treated as the point prediction at $\x$, and the posterior variance gives the associated prediction uncertainty. Regardless of being the best linear unbiased predictor under the assumed model, a major drawback with GP modeling that limits its applicability to Big Data is its $\mathcal{O}(n^3)$ computational complexity, where $n$ is the number of observations in the training data. This is because GP modeling requires the inversion of an $n \times n$ kernel (or correlation) matrix $\Rmat_{nn}$ involved in posterior estimation. Building GP approximations that scale reasonably well with large $n$ is an active area of research, and the various methodologies to do so can be roughly grouped into two categories: $(i)$ global, and $(ii)$ local approximations. 

Global approximations generally $(i)$ focus on a subset of the training data with $m \ll n$ observations resulting in a much smaller $\Rmat_{mm}$ to invert \citep{chalupka2013}, where the subset can be randomly selected, based on clustering, or with active learning \citep{lawrence2002, keerthi2005}, $(ii)$ use a compactly supported kernel function \citep{gneiting2002} to generate sparse $\Rmat_{nn}$, then exploit the sparsity to efficiently compute $\Rmat_{nn}^{-1}$ \citep{Kaufman2011}, or $(iii)$ approximate $\Rmat_{nn}$ with a low $m$-rank matrix plus diagonal using $m \ll n$ inducing points, that can be inverted in $\mathcal{O}(m^2n)$ \citep{candela2005}. Local approximations essentially $(i)$ build a local GP model for each testing location \citep{gramacy2015}, or $(ii)$ partition the training data into disjoint blocks and build independent GP for each block resulting in a block diagonal $\Rmat_{nn}$ that can be inverted efficiently \citep{das2010}.

Global or local approximations alone can be insufficient depending upon the data. To model a complex latent function $f(\x)$ that varies significantly locally, a larger $m$ would be needed to adequately summarize the data, be it subset of data approach or inducing points. Local approximations where a local GP is fit in the neighborhood of the testing location ignore the global trend and often result in over-confident predictions due to local over-fitting. On the other hand, local approximations where independent GPs are fit on partitioned training data suffer from discontinuity at the boundaries of the local regions. \cite{chiwoo2016} show that greater the discontinuity lower the prediction accuracy, especially at the boundaries. 

\cite{snelson2007} build on the inducing points framework presented in \cite{candela2005} to incorporate local trend. To begin with, the inducing points framework makes the assumption that given the inducing points $\mathcal{I}$, the training and testing conditional distributions are independent, i.e, for any training location $\x$ and testing location $\x^*$, $p(f(\x), f(\x^*) | \mathcal{I}) = p(f(\x)| \mathcal{I}) p(f(\x^*) | \mathcal{I})$. The fully independent conditional approximation (\texttt{FIC}) given in \cite{snelson2005} makes additional assumption that given $\mathcal{I}$, the conditional distribution of the latent function at any two locations (training and/or testing) $\x_i, \x_j$ are independent, i.e., $p(f(\x_i), f(\x_j) | \mathcal{I}) = p(f(\x_i) | \mathcal{I}) p(f(\x_j) | \mathcal{I})$. \cite{snelson2007} later present partially independent conditional approximation (\texttt{PIC}) where the space is partitioned into disjoint blocks and given $\mathcal{I}$, the conditional distribution at any two locations are independent only when they belong to separate blocks. The dependence within a block in \texttt{PIC} attempts to capture the local trend, with the limitation that at boundaries of a block the dependence from locations in neighboring blocks are ignored.

The discontinuity problem alluded to before with local approximations that build independent GPs on partitioned training data is generally addressed by using some form of weighted averaging of the independent GPs \citep{tresp2000, rasmussen2001, gramacy2008, chen2009bagging, deisenroth2015}. \cite{park2018patchwork} present a patching of the independent GPs (\texttt{PK}) where they augment the training data with a set of pseudo observations located at the boundaries of neighboring regions, and impose continuity by enforcing two neighboring GPs to make identical predictions at the pseudo locations common to them. Though the discontinuity problem can be accounted for, the global trend remains elusive for local approximations.

In this work we propose a novel methodology to capture both global and local trend in GP modeling. We first select a set of observations from the training data, independent of the testing location, to capture the global trend. The global point set is then supplemented with observations from the neighborhood of a given testing location to capture the local trend. The global trend is modeled using a power exponential kernel whose hyperparameters are learned based on the global points alone, and the local trend is modeled using a compactly supported kernel with a single hyperparameter that is predetermined based on the global points. Both kernels act on the combined set of global and local points of size $m \ll n$, resulting in an additive kernel. \cite{vanhatalo2008} also use an additive kernel where the global trend is modeled with \texttt{FIC} and local trend with a compactly supported kernel, however, contrary to our approach, their compactly supported kernel acts on the full training data and requires the training data to have lattice structure to efficiently invert the kernel matrix by exploiting its sparsity. Our framework does not impose any restrictions on the training data, and scales well in higher dimensions as we do not rely on the sparsity of the compactly supported kernel matrix for efficient inversion. Furthermore, unlike \texttt{FIC} and \texttt{PIC}, we do not make any assumptions on the conditional distributions. 

We refer to our methodology as \texttt{TwinGP} owing to the twin set of training points and kernels involved. The remainder of the article is organized as follows. Section \ref{sec:review} provides a brief review of GP and introduces notation. Section \ref{sec:glgp} presents the new \texttt{TwinGP} framework. Section \ref{sec:ill} gives an illustration of \texttt{TwinGP} using a $1d$ function, and in Section \ref{sec:exp} we compare \texttt{TwinGP} with popular global, and local GP approximations for emulation as well as modeling real world datasets. Finally, in Section \ref{sec:conc} we provide our concluding remarks.

\section{Gaussian Process Review} \label{sec:review}
Denote the training data as $\set{\X_n, \Y_n} \coloneqq \set{\x_i, y_i}_{i=1}^n$ such that the inputs $\x_i \in \R^d$ and output $y_i \in \R$, for all $i$. Let 
\begin{equation}\label{eq:model}
    y_i=f(\x_i)+\epsilon_i, \;\textrm{for} \; i=1,\ldots,n,
\end{equation}
where $\epsilon_i \overset{iid}{\sim} \N(0,\nu^2)$ is the random noise in the output. Our aim is to estimate the latent function $f(\cdot)$ from the training data. In GP modeling, we assume that $f(\cdot)$ to be a realization from a Gaussian process:
\begin{equation}\label{eq:GP}
    f(\x)\sim \text{GP}(\mu, \tau^2 \Rcov(\x, \cdot)),
\end{equation}
where $\mu$ $\tau^2$, and $\Rcov(\cdot,\cdot)$ are the mean, variance, and correlation function of the GP, respectively. The correlation function is defined as $Cor\{f(\bm u),f(\bm v)\} = \Rcov(\bm u,\bm v)$, which is a positive definite function with $\Rcov(\bm u,\bm u)=1$. We use the names correlation function and kernel function interchangeably in this paper. Please refer to the books \cite{santnerbook} and \cite{gramacy2020surrogates} for details on GP modeling. 

From (\ref{eq:model}) and (\ref{eq:GP}), we have $\bm Y_n\sim \N_n(\mu\one_n, \tau^2\Rmat_{nn}+\nu^2\bm I_n)$, where $\Rmat_{nn}$ is the $n\times n$ correlation matrix whose $ij^{th}$ element is $\Rcov(\x_i,\x_j)$,  $\one_n \coloneqq [1, \dots, 1]^{\prime}$, and $\bm I_n$ is the $n\times n$ identity matrix. Let $\eta = \nu^2/\tau^2$, known as nugget. Given an arbitrary testing location $\x^* \in \R^d$, we are interested in finding the conditional distribution of $f(\x^*) |\Y_n$.

Define the $1\times n$ correlation vector as $\Rmat(\x^*, \X_n) \coloneqq [\Rcov(\x^*, \x_1), \dots, \Rcov(\x^*, \x_n)]$. In keeping with the notation, we have 
 $\Rmat(\X_n,\x^*)=\Rmat(\x^*, \X_n)'$ and $\Rmat_{nn} = \Rmat(\X_n, \X_n) \coloneqq [\bm R (\x_1, \X_n); \dots; \Rmat(\x_n, \X_n)]$. The joint distribution of $f(\x^*)$ and $\bm f(\X_n) \coloneqq [f(\x_1), \dots, f(\x_n)]^{\prime}$ is given by
\begin{align*}
    \begin{bmatrix} f(\x^*) \\ \bm f(\X_n) \end{bmatrix} &\sim \N_{n + 1}\Bigg(\begin{bmatrix} \mu \\ \mu\one_n \end{bmatrix},\ \tau^2 \begin{bmatrix} 1 & \Rmat(\x^*, \X_n) \\ \Rmat(\X_n, \x^*) & \Rmat_{nn} + \eta \I_n \end{bmatrix} \Bigg).
\end{align*} 
Then, the conditional distribution of $f(\x^*) | \Y_n$, is given by
\[f(\x^*) | \Y_n\sim \N(\mu(\x^*), \sigma^2(\x^*)),\]
where
\begin{align}
    \mu(\x^*) &= \mu + \Rmat(\x^*, \X_n) [\Rmat_{nn} + \eta \I_n]^{-1} (\Y_n - \mu\one_n), \label{eq:mu} \\ 
    \sigma^2(\x^*) &= \tau^2 \Big\{ 1 - \Rmat(\x^*, \X_n) [\Rmat_{nn} + \eta \I_n]^{-1} \Rmat(\X_n, \x^*) \Big\}. \label{eq:sigma}
\end{align}

There exists a multitude of correlation functions in the literature \citep[Ch.4]{rasmussenbook}, each with their own set of correlation parameters $\btheta$ that are estimated from the training data. The empirical Bayes estimate of $\mu$, $\tau^2$, $\eta$, and $\btheta$ are given as follows \citep{santnerbook}:
\begin{align}
    \hat{\mu} &= \frac{\one_n^{\prime} [\Rmat_{nn} + \eta \I_n]^{-1} \Y_n}{\one_n^{\prime}\ [\Rmat_{nn} + \eta \I_n]^{-1}\ \one_n}, \label{eq:muhat} \\
    \hat{\tau}^2 &= \frac{1}{n} (\Y_n - \hat{\mu}\one_n)^{\prime} [\Rmat_{nn} + \eta \I_n]^{-1} (\Y_n - \hat{\mu}\one_n), \label{eq:tauhat} \\
    \hat{\btheta}, \hat{\eta} &= \argmin_{\btheta,\ \eta\ >\ 0}\ n\log{\hat{\tau}^2} + \log|\Rmat_{nn} + \eta \I_n|. \label{eq:thetahat}
\end{align}
As evident from the above equations, much of the complexity in GP modeling stems from the computation of $[\Rmat_{nn} + \eta \I_n]^{-1}$ and $|\Rmat_{nn} + \eta \I_n|$, both requiring $\mathcal{O}(n^3)$ operations. In the next section, we propose our methodology to address this computational bottleneck for large $n$.

\section{Global-Local Gaussian Process} \label{sec:glgp}
In order to circumvent the $\mathcal{O}(n^3)$ computational complexity, we build the GP using only $m \ll n$ points form the training data. Contrary to the global approximation methodologies in the literature where the $m$ points are chosen from the training data or artificially generated, so as to summarize the whole data, we use a combination of $g$ global points and $l$ local points such that $m = g + l$. The $g$ global points are independent of the testing location, while the $l$ local points are specific to the testing location $\x^*$.

To capture the global trend, and local trend with respect to $\x^*$, the kernel function $\Rcov(\cdot,\cdot)$ is also modeled as a combination of two kernel functions \citep{ba2012composite, harari2014convex}, i.e., 
\begin{align}
    \Rcov(\x_a, \x_b) &= (1 - \lambda) \Gcov(\x_a, \x_b) + \lambda \Lcov(\x_a, \x_b),\ \lambda \in [0, 1]. \label{eq:cov}
\end{align}
$\Gcov$ is designed to capture the global trend in the data, while $\Lcov$ captures the local trend around $\x^*$. The hyperparameter $\lambda$ controls the proportion of global and local trends used in building $\Rcov$.

Let $\X_m \subset \X_n$ represent the $m$ training points, then $\Gmat_{mm} = \Gcov(\X_m, \X_m)$, and $\Lmat_{mm} = \Lcov(\X_m, \X_m)$. With this setup, instead of inverting $\Rmat_{nn} + \eta \I_n$ in (\ref{eq:mu})-(\ref{eq:thetahat}), we need only invert $\Rmat_{mm} + \eta \I_m$ where
\begin{align}
    \Rmat_{mm} &= (1 - \lambda)\Gmat_{mm} + \lambda \Lmat_{mm}. 
\end{align}
Thus, the computational complexity of fitting the GP reduces to $\mathcal{O}(m^3)$.  In the subsections that follow, we present how to efficiently locate the $m$ training points, the final predictive equations, and how the kernel function $\Rcov(\cdot,\cdot)$ in (\ref{eq:cov}) is determined.

\subsection{Global and Local Points} \label{subsec:train}
Denote the global training points as $\X_g$, and the local training points as $ \X_l$, we have $\X_m = \X_g \cup \X_l$. Ideally, for a given testing location $\x^*$, we should be denoting $\X_m$ as $\X_m(\x^*) = \X_g \cup \X_l(\x^*)$, a technicality we omit for ease of notation.

The objective we desire to achieve with $\X_g$ is to sufficiently model the global trend in the training data. \cite{Mak2018} proposed a nonparametric and model-free data reduction technique known as support points that can produce a small set of points to represent the full dataset. Support points are obtained by minimizing the energy distance \citep{szekely2013energy} between its empirical distribution and that of the dataset using difference-of-convex programming. However, support points need not be a subset of the full dataset. Given the support points, \cite{joseph2021split} use sequential nearest neighbor assignment to sample from the dataset. Even still, computing support points as given in \cite{Mak2018} is $\mathcal{O}(n^2)$. Later, \cite{twinning} present an efficient sampling algorithm named \texttt{Twinning} that minimizes the energy distance in $\mathcal{O}(n \log{n})$. Since we are dealing with large datasets in this work, we will use \texttt{Twinning} to find $\X_g$.

\begin{figure}[ht]
\begin{center}
\includegraphics[width = 0.75\textwidth]{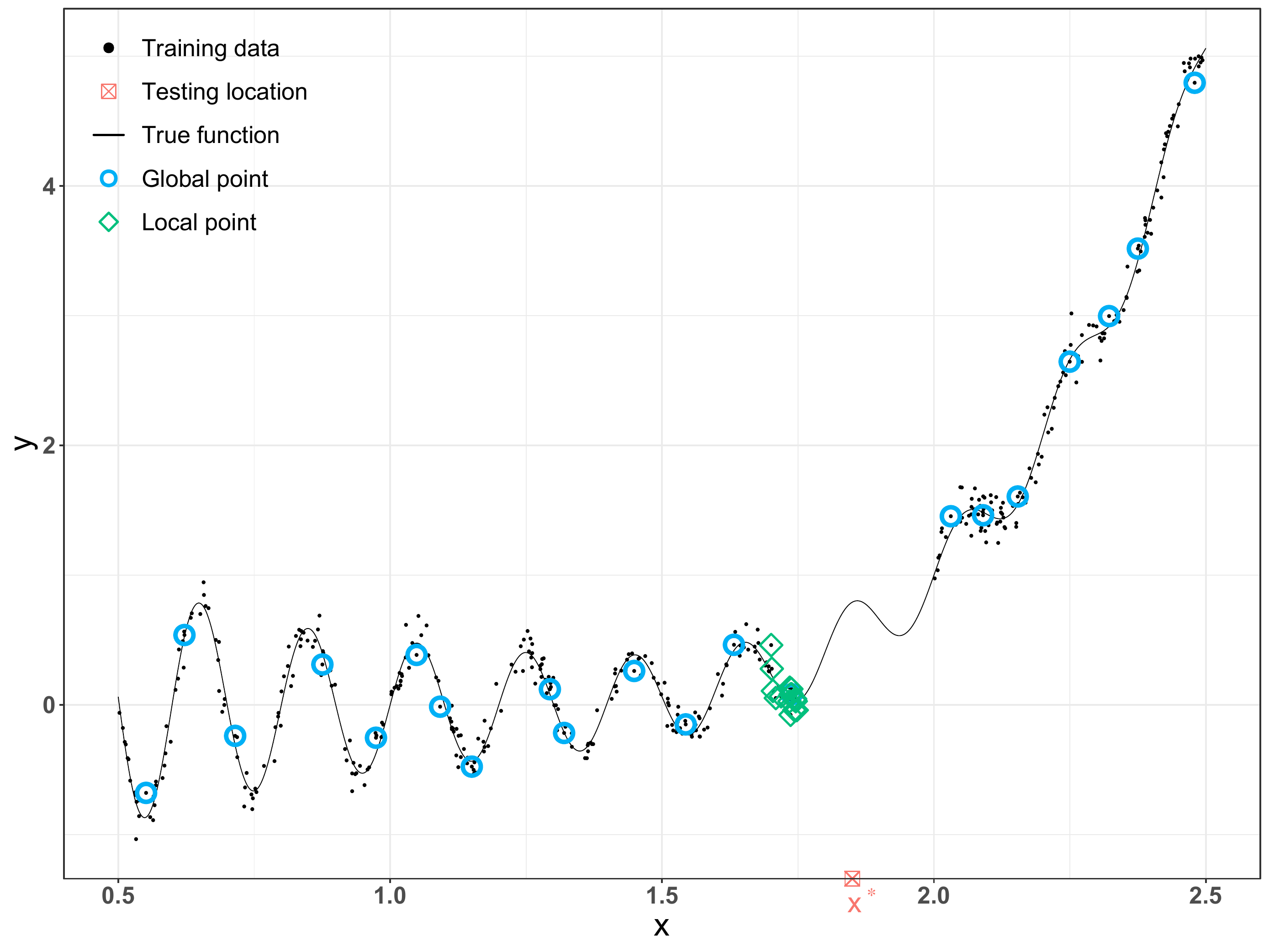} 
\caption{Illustration of the training points (global and local) identified for a given training data and testing location with \texttt{TwinGP}.}
\label{fig:1d_points}
\end{center}
\end{figure}

Given the testing location $\x^*$, $\X_l$ is obtained as the $l$ nearest neighbors to $\x^*$ in $\X_n \setminus \X_g$, a task that can be performed efficiently with $kd$-tree in $\mathcal{O}(l \log{l})$. Figure \ref{fig:1d_points} gives a depiction of the global and local training points identified for a given testing location, for a $1d$ example. One can observe from the figure that fitting a GP on the local points alone, shown as green diamonds, will clearly underpredict at the testing location, and it is shown to be not optimal \citep{emery2009kriging}. In \texttt{laGP} proposed by \cite{gramacy2015}, the nearest neighbors are supplemented with active learning, however, an optimization is required at each testing location that adds to the computational complexity. On the other hand, our method makes use of the predetermined global points (blue circles) to capture the global trend, and therefore, nearest neighbors alone as the local points will suffice. This reduces the computational burden in our framework.

\subsection{Predictive equations} \label{subsec:pred}
At this stage, given $\x^*$, we have identified $\X_m$. Assume that the kernel function $\Rcov$ is fully determined. Let $\Y_m$ be the response vector corresponding to $\X_m$, i.e., $\Y_m = [\Y_g; \Y_l]$ where $\Y_g$ and $\Y_l$ are the response vectors corresponding to $\X_g$ and $\X_l$, respectively. The conditional distribution of $f(\x^*)|\Y_m$ is given by $\N(\mu(\x^*), \sigma^2(\x^*))$, where
\begin{align}
    \mu(\x^*) &= \hat{\mu}_m + \Rmat(\x^*, \X_m) [\Rmat_{mm} + \eta \I_m]^{-1} (\Y_m - \mu\one_m), \label{eq:finalmu} \\ 
    \sigma^2(\x^*) &= \hat{\tau}^2_m \Big\{ 1 - \Rmat(\x^*, \X_m) [\Rmat_{mm} + \eta \I_m]^{-1} \Rmat(\X_m, \x^*) \Big\}, \label{eq:finalsigma} \\
    \hat{\mu}_m &= \frac{\one_m^{\prime} [\Rmat_{mm} + \eta \I_m]^{-1} \Y_m}{\one_m^{\prime} [\Rmat_{mm} + \eta \I_m]^{-1} \one_m}, \\
    \hat{\tau}^2_m &= \frac{1}{m} (\Y_m - \hat{\mu}_m\one_m)^{\prime} [\Rmat_{mm} + \eta \I_m]^{-1} (\Y_m - \hat{\mu}_m\one_m).
\end{align}
For the prediction of a noisy observation, we have  $y^*|\Y_m \sim \N(\mu(\x^*), \sigma_{\eta}^2(\x^*))$, where $\mu(\x^*)$ is the same as given in (\ref{eq:finalmu}), but $ \sigma_{\eta}^2(\x^*)$ changes to
\begin{align}
    \sigma_{\eta}^2(\x^*) &= \hat{\tau}^2_m \Big\{ 1 + \eta - \Rmat(\x^*, \X_m) [\Rmat_{mm} + \eta \I_m]^{-1} \Rmat(\X_m, \x^*) \Big\}. \label{eq:finalsigma_noise} 
\end{align}
Furthermore, $\Rmat_{mm} + \eta \I_m$ can be deconstructed as follows,
\begin{align}
    \Rmat_{mm} + \eta \I_m &= \begin{bmatrix} \Rmat_{gg} + \eta \I_g & \Rmat_{gl} \\ \Rmat_{lg} & \Rmat_{ll} + \eta \I_l \end{bmatrix} \\
    &= \begin{bmatrix} (1 - \lambda) \Gmat_{gg} + \lambda \Lmat_{gg}+\eta\bm I_g & (1 - \lambda) \Gmat_{gl} + \lambda \Lmat_{gl} \\ (1 - \lambda) \Gmat_{lg} + \lambda \Lmat_{lg} & (1 - \lambda) \Gmat_{ll} + \lambda \Lmat_{ll}+\eta\bm I_l \end{bmatrix}. \label{eq:block_mat}
\end{align}
Since $\Rmat_{gg} = (1 - \lambda) \Gmat_{gg} + \lambda \Lmat_{gg}$ is independent of the testing location $\x^*$, $\Rmat_{gg}$ and $[\Rmat_{gg} + \eta \I_g]^{-1}$ can be predetermined before testing. For a given $\x^*$, the computational effort in inverting $\Rmat_{mm} + \eta \I_m$ can be significantly reduced by using block matrix inversion as follows,
\begin{align}
    [\Rmat_{mm} + \eta \I_m]^{-1} &= \begin{bmatrix} \Sig_{gg}^{-1} +  \Sig_{gg}^{-1} \Rmat_{gl} \bm{S}^{-1} \Rmat_{lg} \Sig_{gg}^{-1} & -\Sig_{gg}^{-1} \Rmat_{gl} \bm{S}^{-1} \\ -\bm{S}^{-1} \Rmat_{lg} \Sig_{gg}^{-1} & \bm{S}^{-1} \end{bmatrix}, \label{eq:block}
\end{align}
where $\Sig_{gg} = \Rmat_{gg} + \eta \I_g$, and $\bm{S} = \Rmat_{ll} - \Rmat_{lg} \Sig_{gg}^{-1} \Rmat_{gl}$, thereby, per testing location we need only invert $\bm{S}$, a small $l \times l$ matrix.

\subsection{Correlation Functions} \label{subsec:global}
To model the global trend, we use the popular power exponential kernel function \citep{sacks1989design},
\begin{align}
    \Gcov(\x_a, \x_b) &= \exp \left( -\sum_{i=1}^d \frac{|\x_a^i - \x_b^i|^{\alpha}}{\theta^i_g} \right),\ \alpha \in (0, 2], \theta_g^i>0.
\end{align}
Denote the hyperparameters of $\Gcov$ as $\btheta_g$, we have $\btheta_g = \set{\theta_g^1, \dots, \theta_g^d, \alpha}$. They can be estimated with respect to $\X_g$ alone, i.e., independent of the testing location,
\begin{align}
    \hat{\mu}_g &= \frac{\one_g^{\prime} [\Gmat_{gg}+\eta_g \bm I_g]^{-1} \Y_g}{\one_g^{\prime} [\Gmat_{gg}+\eta_g \bm I_g]^{-1} \one_g}, \\
    \hat{\tau}^2_g &= \frac{1}{g} (\Y_g - \hat{\mu}_g\one_g)^{\prime} [\Gmat_{gg}+\eta_g \bm I_g]^{-1} (\Y_g - \hat{\mu}_g\one_g), \\
    \hat{\btheta}_g &= \argmin_{\btheta_g}\ g\log{\hat{\tau}^2_g} + \log|\Gmat_{gg}+\eta_g \bm I_g|.\label{eq:thetag}
\end{align}
We have constrained $\alpha\in [1,2]$ in the optimization assuming the latent function $f(\cdot)$ to be reasonably smooth and not too wiggly.

To model the local trend, we use compactly supported correlation functions \citep{gneiting2002}, which ensures identifiability of the local correlation parameters with respect to the global correlation parameters. Specifically, we use Wendlands's compactly supported radial function \citep{wendland_2004},
\begin{align}
    \Lcov(\x_a, \x_b) &= \left((q + 1) \frac{\norm{\x_a - \x_b}_2}{\theta_l} + 1 \right) \max \left\{0, \left(1 - \frac{\norm{\x_a - \x_b}_2}{\theta_l} \right) \right\}^{q + 1},\ q = \lfloor \frac{d}{2} \rfloor + 2
\end{align}
where  $\norm{\cdot}_2$ is the $\ell_2$ norm and $\lfloor u \rfloor$ is the largest integer lesser than or equal to $u$. $\Lcov$ has a single hyperparameter $\theta_l$ which we set as the covering radius \citep[p. 22]{fasshauer2007meshfree} for $\X_g$ to cover $\X_n$, i.e., 
\begin{align}
    \hat{\theta}_l &= \min \left\{\rho: \X_n \subseteq \cup_{i=1}^g \mathcal{B}_{\rho}(\x_i),\ \x_i \in \X_g,\ \forall i \right\}, \label{eq:thetal}
\end{align}
where $\mathcal{B}_{\rho}(\x)$ is a closed ball of radius $\rho$ centered at $\x$.  Setting $\theta_l$ as in (\ref{eq:thetal}) makes it independent of the testing location, thereby, allowing us to precompute $\Lmat_{gg}$ in (\ref{eq:block_mat}) leading to the computational gains with block matrix inversion (\ref{eq:block}). In addition, almost surely for any testing location $\x^*$ there exists at least one global training point such that the local kernel is active between them, i.e., there exists $\x_g \in \X_g: \Lcov(\x^*, \x_g) > 0$. The motivation here is that we do not wish $\Lcov$ to neglect the correlation between $\x^*$ and $\X_g$.

We need to specify two more parameters: $\lambda$ and $\eta_l$. We could estimate them using empirical Bayes methods as before, but since we have unused data in the training set, we can do the estimation in a different and more robust fashion. We sample $\set{\X_v, \Y_v}$ from $\set{\X_n, \Y_n} \setminus \set{\X_g, \Y_g}$ by \texttt{Twinning} to create a validation set. Now, we can estimate $\lambda$ and $\eta_l$ by minimizing the prediction error:
\begin{align}
    \hat{\lambda},\ \hat{\eta_l} &= \argmin_{\lambda \in [0, 1],\ \eta_l > 0}\ \text{MSE}(\lambda) \label{eq:lamnoiseopt} \\
    &= \argmin_{\lambda \in [0, 1],\ \eta_l > 0}\ \sum_{\x \in \X_v} (y_{\x} - \mu(\x))^2, \nonumber
\end{align}
where $\mu(\x)$ is obtained as given in (\ref{eq:finalmu}) with $\eta = (1 - \lambda) \eta_g + \lambda \eta_l$.

\subsection{Computational complexity}
The \texttt{TwinGP} procedure is summarized in Algorithm \ref{alg:algo}. The computational complexity of \texttt{TwinGP} can be deconstructed as follows,
\subsubsection*{Testing}
\begin{itemize}
    \item[(i)] Identifying $\X_l$ for a given $\x^*$ is on average $\mathcal{O}(\log{n})$ using $kd$-tree.
    \item[(ii)] The complexity in computing $\mu(\x^*)$ and $\sigma^2(\x^*)$ is dictated by inversion of $\Rmat_{mm}$. With block matrix inversion as given in (\ref{eq:block}), computing $[\Rmat_{mm} + \eta \I_m]^{-1}$ is $\mathcal{O}(g^2l)$.
\end{itemize}

\subsubsection*{Training}
\begin{itemize}
    \item[(i)] Obtaining a twinning sample to identify $\X_g$ is on average $\mathcal{O}(n \log{n})$.
    \item[(ii)] The complexity in estimating $\btheta_g$ is dictated by inversion of $\Gmat_{gg} + \eta_g \I_g$ that is $\mathcal{O}(g^3)$.
    \item[(iii)] $\theta_l$ can be estimated efficiently with a $kd$-tree in $\mathcal{O}(n \log{g})$.
    \item[(iv)] The complexity in estimating $\lambda$ and $\eta_l$ is similar to testing, i.e., $\mathcal{O}(g^2l)$ per validation point.
\end{itemize}

\subsubsection*{Overall complexity}
Thus, the overall computational complexity of \texttt{TwinGP} is $\mathcal{O}(g^3 + tg^2l)$, where $t$ is the number of testing locations. If we select $g$ to be order of $\sqrt{n}$, the training complexity reduces to $\mathcal{O}(n^{1.5})$, a substantial improvement over $\mathcal{O}(n^3)$.

\begin{algorithm}
\caption{\texttt{TwinGP}}
\label{alg:algo}
\begin{algorithmic}[1]
\State Input training set $\set{\X_n, \Y_n}$ and testing locations $\X^*_t$
\State Identify global training points $\X_g$ (Section \ref{subsec:train})
\State Estimate $\btheta_g$ as in (\ref{eq:thetag})
\State Estimate $\theta_l$ as in (\ref{eq:thetal})
\State Estimate $\lambda$ and $\eta_l$ as in (\ref{eq:lamnoiseopt})
\For{$\x^* \in \X^*_t$}
    \State Identify local training points $\X_l$ (Section \ref{subsec:train})
    \State Compute $\mu(\x^*)$ and $\sigma^2(\x^*)$ (Section \ref{subsec:pred})
\EndFor
\State \textbf{end for}
\State \textbf{return} $\set{\mu(\x^*), \sigma^2(\x^*)}, \forall \x^* \in \X^*_t$
\end{algorithmic}
\end{algorithm}

\section{$1d$ Illustration} \label{sec:ill}
\begin{figure}[ht]
\begin{center}
\includegraphics[width = \textwidth]{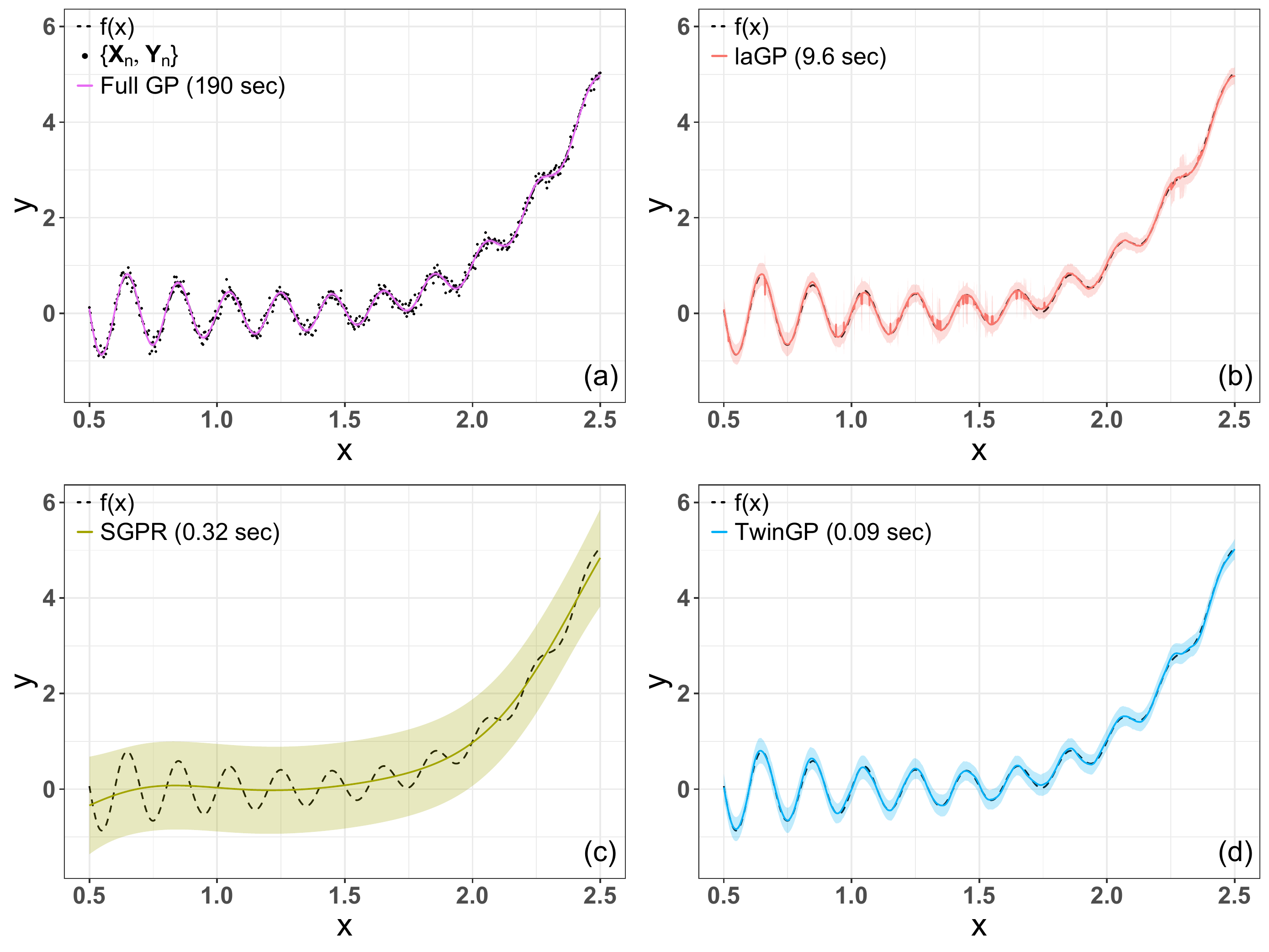} 
\caption{Prediction and $2\sigma$ confidence interval (shaded region) with full GP, \texttt{laGP}, \texttt{SGPR}, and \texttt{TwinGP} to model the $1d$ function in (\ref{eq:1d}).}
\label{fig:1d}
\end{center}
\end{figure}

Here we illustrate how \texttt{TwinGP} overcomes the shortcomings of purely local or global GP approximations. Consider the following function from \cite{gramacy2012cases},
\begin{align}
    f(x) &= \frac{\sin{10 \pi x}}{2x} + (x - 1)^4,\ x \in [0.5, 2.5]. \label{eq:1d}
\end{align}
We first generate the training dataset  with $n=500$ observations, where $x_1, \dots, x_n$ are selected on a uniform grid in $[0.5, 2.5]$, and we add a Gaussian noise to each observation as follows,
\begin{align}
    y_i &= f(x_i) + \epsilon_i,\ \epsilon_i \sim \N(0, 0.01),\ \forall i = 1, \dots, 500.
\end{align}
The testing set at $2,000$ locations are also selected on a uniform grid in $[0.5, 2.5]$. Plot (a) in Figure \ref{fig:1d} depicts $f(x)$, the 500 training locations, and the prediction from a full GP modeled using the \texttt{mlegp} \citep{mlegp} package. As evident, $f(x)$ is recovered extremely well with a full GP, in addition, the confidence intervals are tight.

To demonstrate the global GP approximation method, we use \texttt{SGPR} \citep{titsias09a} implemented in \texttt{GPyTorch} \citep{gardner2018gpytorch}. \texttt{SGPR} is a low-rank GP approximation where the inducing points and kernel hyperparameters are estimated with a variational learning approach. For local approximation we use \texttt{laGP} \citep{gramacy2015} implemented in the \texttt{laGP} \citep{lagp_package} package. For each testing location, \texttt{laGP} starts with a set of training points in its neighborhood and then sequentially adds more training points so as to minimize the predictive variance.

For \texttt{TwinGP} we set the number of global points $g=22$ and local points $l=25$. For \texttt{SGPR} the number of inducing points is set to be $m = g + l$, and for \texttt{laGP} $l + 10$ training locations are considered per testing location, i.e., start with $l$ neighborhood points and then sequentially add 10 more points. Plots (b), (c), and (d) in Figure \ref{fig:1d} give the predictions and $2\sigma$ confidence intervals from \texttt{laGP}, \texttt{SGPR}, and \texttt{TwinGP} respectively. As expected, \texttt{laGP} predictions are discontinuous owing to its local nature, while \texttt{SGPR} produces smooth prediction neglecting the local oscillations of $f(x)$. On the other hand, \texttt{TwinGP} gives comparable predictions to the full GP. The total computation time for training and testing with the four methods are given in Figure \ref{fig:1d}. We can clearly see the computational saving with \texttt{TwinGP} over the full GP, which will be even more substantial with large datasets as we demonstrate in the next section.

\section{Experiments} \label{sec:exp}
In this section, we make an extensive analysis of \texttt{TwinGP} performance on several emulation, and real world datasets, compared to other scalable GP modeling frameworks. Similar to Section \ref{sec:ill}, in our experiments we include \texttt{SGPR} for global approximation, \texttt{laGP} for local approximation, and patchwork kriging (\texttt{PK}) by \cite{park2018patchwork} introduced in Section \ref{sec:intro}. All the experiments presented in this section are carried out on a $2.6$ GHz 6-Core Intel i7-9750H processor with 16 GB memory. We apply the following general settings for the different modeling frameworks, unless stated otherwise.
\begin{enumerate}
    \item[\texttt{TwinGP}:] The number of global points $g$ is set as $\min\set{50d,\ \max\set{\lfloor \sqrt{n} \rfloor,\ 10d}}$, i.e., at least $10d$ points are chosen to model the global trend, with an upper bound of $50d$. The local trend is modeled with $l = \max\set{25,\ 3d}$ points. The number of validation points used for estimating $\lambda$ and $\eta$ is set as $2g$. The $\btheta_g$ hyperparameter optimization is done with grid initialization and multi-starts as outlined in \cite{basak2022numerical}.
    \item[\texttt{SGPR}:] For a fair comparison with $\texttt{TwinGP}$, we set the number of inducing points to be $m = g + l$. We follow the implementation provided in \texttt{GPyTorch} documentation\footnote{\url{https://docs.gpytorch.ai/en/latest/examples/02_Scalable_Exact_GPs/SGPR_Regression_CUDA.html}}. The number of iterations in hyperparameter optimization is modified from constant 100 to $\min\set{250,\ \lfloor 50 \log{(1 + d)} \rfloor}$, and a learning rate of $0.05$ is used instead of $0.01$. In addition, separate lengthscales are learnt per dimension.
    \item[\texttt{laGP}:] For each testing location, we start with $l$ training points in its neighborhood and sequentially add 10 more points to the design that minimize the predictive variance. The \texttt{aGPsep} function provided in the \texttt{R} package by \cite{lagp_package} is used to execute \texttt{laGP} in parallel. For emulation datasets where there is no observation noise, the nugget parameter is set to $10^{-7}$, and for real world datasets the nugget is set to \texttt{NULL} in which case it is estimated.
    \item[\texttt{PK}:] The training locations are partitioned into $K$ disjoint blocks such that each block contains at least $10d$ points. The spatial tree algorithm used to make the partition benefits from $K$ being a power of 2, hence, we set $K = 2^{\lfloor \log_2{\frac{n}{10d}} \rfloor}$. The number of pseudo observations $B$ introduced at the boundaries of neigboring blocks to enforce continuity between the neighboring GPs is set to be 3. Our choice of $B$ is conservative as the complexity of \texttt{PK} scales proportionally to $B^3$. We use the \texttt{MATLAB} implementation of \texttt{PK} provided by the author\footnote{\url{https://www.chiwoopark.net/code-and-dataset}}.
\end{enumerate}

\subsection{Evaluation criteria} \label{sec:eval}
The performance of the different modeling frameworks is assessed based on their prediction accuracy and total computation time. The prediction accuracy is quantified with root mean squared error (RMSE) and negative log predictive density (NLPD). RMSE measures the quality of point predictions from the model, while NLPD measures how well the predicted posterior distribution fits the testing data. Lower the RMSE and NLPD values, better the performance of the model. Given the testing set $\set{\X^*_t, \Y^*_t}$ with $t$ testing locations, we have
\begin{align}
    \text{RMSE} &= \sqrt{\frac{1}{t} \sum_{i=1}^t \{y^*_i- \mu(\x^*_i) \}^2}, \\
    \text{NLPD} &= \frac{1}{2t} \sum_{i=1}^t \left[ \frac{\{y^*_i- \mu(\x^*_i) \}^2}{\sigma^2(\x^*_i)} + \log\{2 \pi \sigma^2(\x^*_i) \} \right].
\end{align}

\subsection{Emulation} \label{subsec:emul}
We consider three emulation problems here, the piston simulation function \citep{kenett2021modern}, the borehole function \citep{morris1993bayesian}, and the Dette \& Pepelyshev function \citep{dette2010generalized}. For a given emulation experiment, the input dimensions are scaled to $[0, 1]$ and $n=10,000$ training locations are sampled uniformly from a $d$-dimensional hypercube, i.e., $\X_n \sim \text{Unif}(0, 1)^{n \times d}$, and $2,000$ deterministic Sobol sequence in $d$ dimensions are the testing locations. A GP is modeled on the training locations using \texttt{TwinGP}, \texttt{SGPR}, \texttt{laGP}, and \texttt{PK}. The performance of the fitted models is assessed based on the criteria given in Section \ref{sec:eval}. The procedure is repeated for 50 iterations with different training locations sampled similary, while the testing locations remain unchanged. The distribution of RMSE and NLPD over the 50 iterations are presented as box-and-whisker plots. Further information and code for the emulations problems can be obtained from Simon Fraser University's virtual library of simulation experiments\footnote{\url{https://www.sfu.ca/~ssurjano/emulat.html}}.

\subsubsection{Piston simulation function}
\begin{figure}[ht]
\begin{center}
\includegraphics[width = \textwidth]{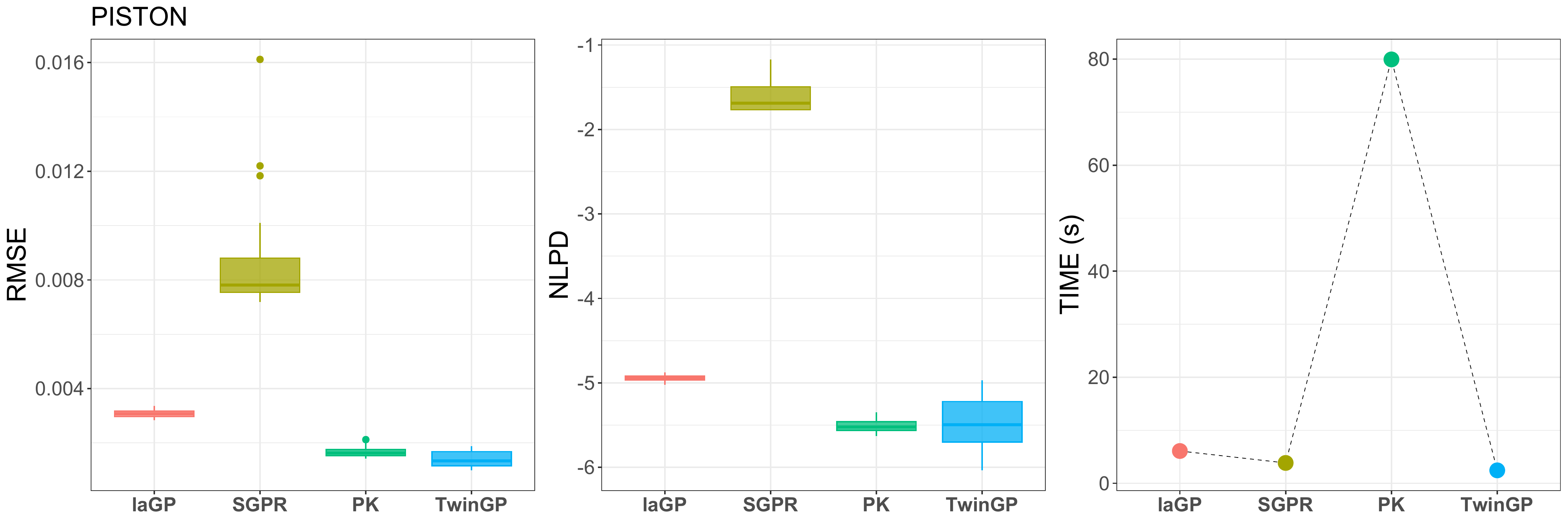} 
\caption{Distribution of RMSE and NLPD over 50 iterations of modeling the piston simulation function with \texttt{laGP}, \texttt{SGPR}, \texttt{PK}, and \texttt{TwinGP}. The corresponding average computation time per iteration is given in the right most plot.}
\label{fig:piston}
\end{center}
\end{figure}

The piston simulation funtion models the cirular motion of a piston within a cylinder \citep{kenett2021modern}. There are $d=7$ inputs, which are piston weight, piston surface area, initial gas volume, spring coefficient, atmospheric pressure, ambient temperature, and filling gas temperature. The response is the time taken to complete one cycle in seconds. The experiment results are given in Figure \ref{fig:piston}. We see that \texttt{TwinGP} performs the best in terms of RMSE, and for NLPD, on average, its performance is comparable to that of \texttt{PK}. Compuation time is the least for \texttt{TwinGP}, around 2 seconds, while \texttt{PK} took around 80 seconds.

\subsubsection{Borehole function}
\begin{figure}[ht]
\begin{center}
\includegraphics[width = \textwidth]{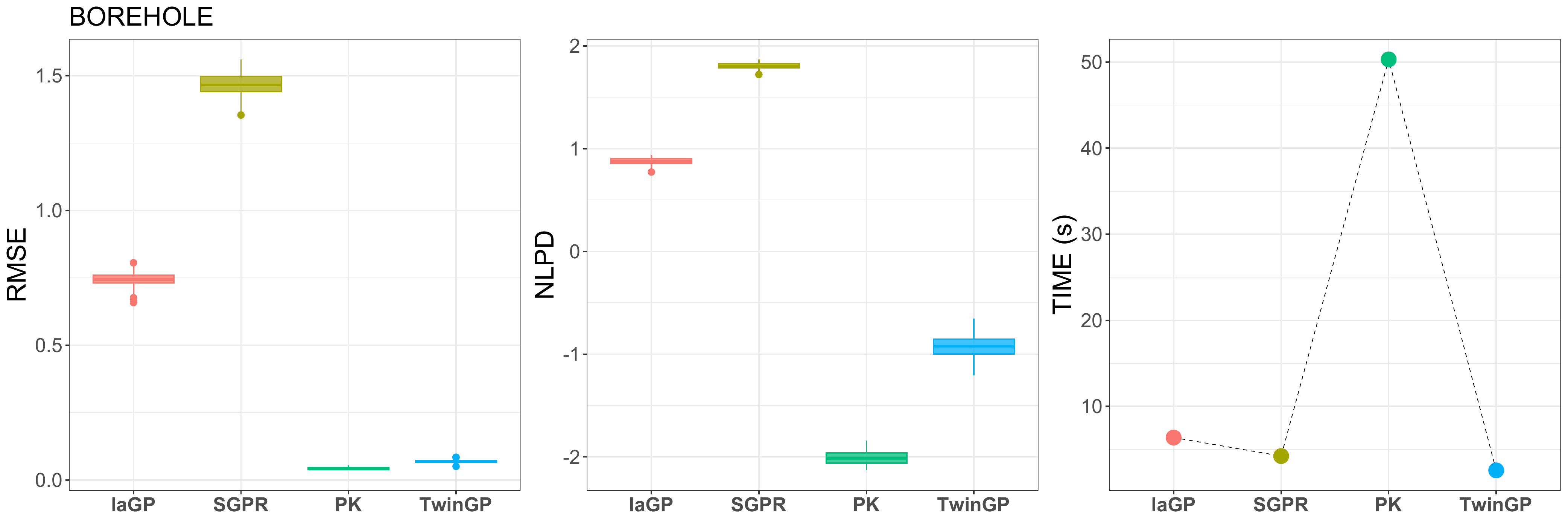} 
\caption{Distribution of RMSE and NLPD over 50 iterations of modeling the borehole function with \texttt{laGP}, \texttt{SGPR}, \texttt{PK}, and \texttt{TwinGP}. The corresponding average computation time per iteration is given in the right most plot.}
\label{fig:borehole}
\end{center}
\end{figure}

The borehole function models water flow through a borehole \citep{morris1993bayesian}. There are $d=8$ inputs, which are radius of the borehole, radius of influence, transmissivity of upper aquifer, potentiometric head of upper aquifer, transmissivity of lower aquifer, potentiometric head of lower aquifer, length of borehole, and hydraulic conductivity of borehole. The response is the water flow rate in $\text{m}^3 / \text{yr}$. The experiment results are given in Figure \ref{fig:borehole}. We see that \texttt{PK} is the best performing modeling framework in terms of both RMSE and NLPD, with \texttt{TwinGP} being the close second. However, in terms of compuation time, \texttt{TwinGP} performance is far superior.

\subsubsection{Dette-Pepelyshev function}
\begin{figure}[ht]
\begin{center}
\includegraphics[width = \textwidth]{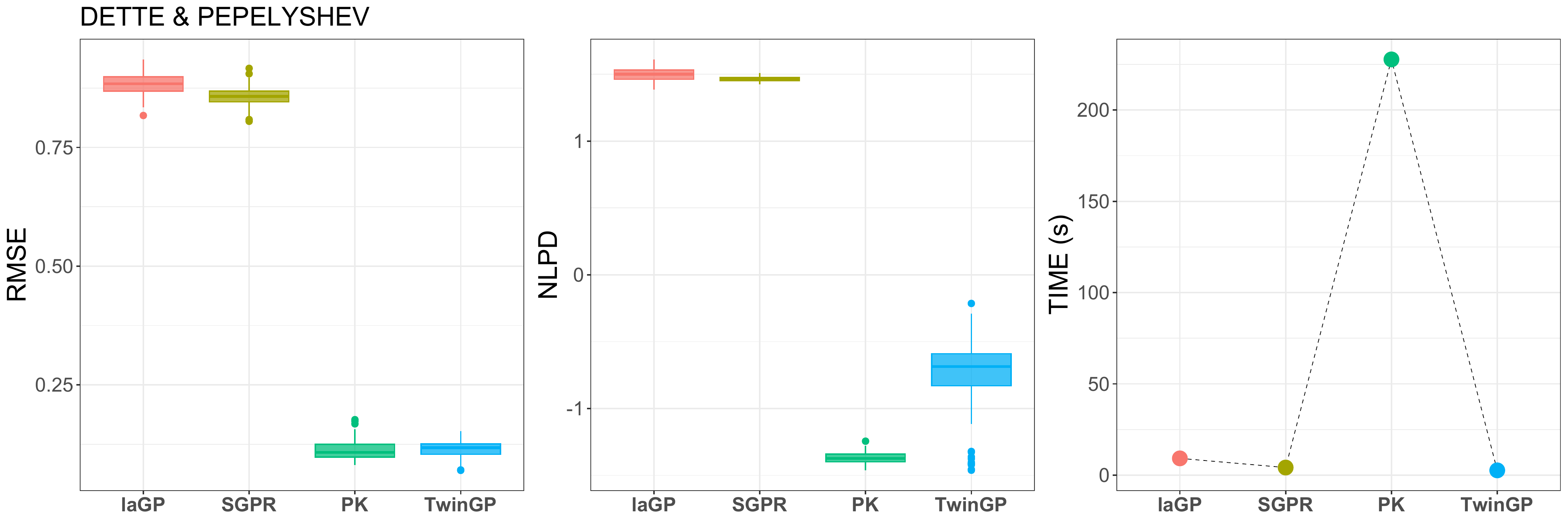} 
\caption{Distribution of RMSE and NLPD over 50 iterations of modeling the Dette-Pepelyshev function with \texttt{laGP}, \texttt{SGPR}, \texttt{PK}, and \texttt{TwinGP}. The corresponding average computation time per iteration is given in the right most plot.}
\label{fig:detpep}
\end{center}
\end{figure}

The Dette-Pepelyshev function from \cite{dette2010generalized} with $d=8$ input variables is heavily curved in some variables, and less so in others. Following the general settings described in the beginning of Section \ref{sec:exp}, running \texttt{PK} with $K=64$ encountered numerical instabilites, hence, we increased $K$ to 128. The experiment results are given in figure \ref{fig:detpep}. In terms of RMSE, both \texttt{TwinGP} and \texttt{PK} perform the best with comparable results. \texttt{PK} performance is the best when it comes to NLPD, though at a very high computational cost. 

\subsection{Real world data} \label{subsec:real}
We consider four real world datasets here: the ozone column spatial dataset, protein tertiary structure dataset, Sarcos robotics dataset, and the flight delays dataset. The same datasets are considered in \cite{park2018patchwork}. For a given dataset, we first randomly split the dataset in 90-10 proportion and a GP is modeled with \texttt{TwinGP}, \texttt{SGPR}, \texttt{laGP}, and \texttt{PK} on $90\%$ of the dataset, and the remaining $10\%$ is used for testing the performance of the fitted models. The experiment is repeated for 50 iterations using different random splits. The distribution of RMSE and NLPD over the 50 iterations are presented as box-and-whisker plots.

\subsubsection{Ozone column}
\begin{figure}[ht]
\begin{center}
\includegraphics[width = \textwidth]{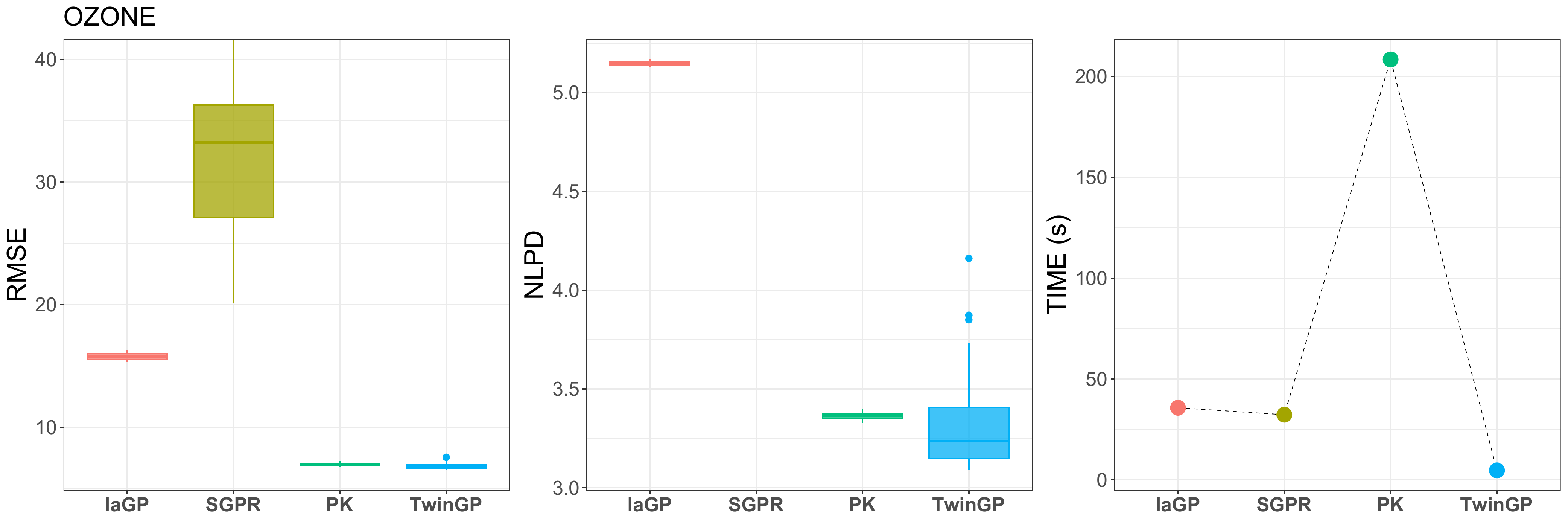} 
\caption{Distribution of RMSE and NLPD over 50 iterations of modeling the ozone column dataset with \texttt{laGP}, \texttt{SGPR}, \texttt{PK}, and \texttt{TwinGP}. The corresponding average computation time per iteration is given in the right most plot.}
\label{fig:ozone}
\end{center}
\end{figure}

The ozone column dataset contains measurement of total column of ozone by the NIMBUS-7/TOMS satellite on October 1, 1988, at different latitudes and longitudes over the globe. There are $182,591$ observations with $d=2$ inputs, latitude and longitude. The response is the total column of ozone. The experiment results are given in Figure \ref{fig:ozone}. We omit \texttt{SGPR} in NLPD plot since it produced negative variances. In terms of RMSE, both \texttt{TwinGP} and \texttt{PK} provide the best performance. For NLPD, \texttt{TwinGP} on average performs better than \texttt{PK}, though \texttt{PK} is more consistent. Furthermore, \texttt{TwinGP} computation time is only 5 seconds, and is the fastest compared to the rest, while \texttt{PK} is the slowest at around 208 seconds. We used $K=1024$ and $B=3$ for \texttt{PK}, following the settings described at the beginning of Section \ref{sec:exp}. Decreasing $K$ to 512 reduced the computation time to about 150 seconds with near identical RMSE and NLPD performance, even still, \texttt{PK} remained the slowest amongst rest of the models. Further decreasing $K$ or increasing $B$ for \texttt{PK} was computationally more expensive with no significant difference in RMSE and NLPD.

\subsubsection{Protein tertiary structure}
\begin{figure}[ht]
\begin{center}
\includegraphics[width = \textwidth]{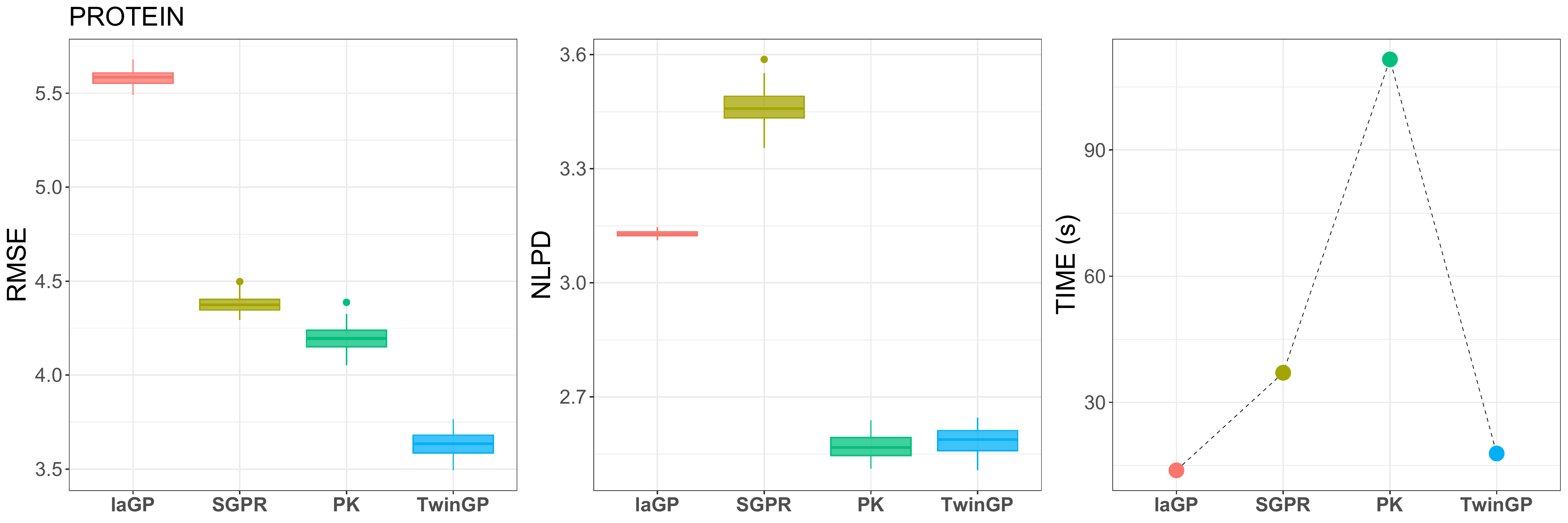} 
\caption{Distribution of RMSE and NLPD over 50 iterations of modeling the protein structure dataset with \texttt{laGP}, \texttt{SGPR}, \texttt{PK}, and \texttt{TwinGP}. The corresponding average computation time per iteration is given in the right most plot.}
\label{fig:protein}
\end{center}
\end{figure}

The protein tertiary structure dataset can be obtained from the UCI machine learning repository\footnote{\url{https://archive.ics.uci.edu/ml/datasets}}. The dataset has $d=9$ input variables relating to the physiochemical properties of protein tertiary structure, and the response is size of the protein residue. There are $45,730$ observations in total. The experiment results are given in Figure \ref{fig:protein}. \texttt{TwinGP} is the best performing model in terms of RMSE. For NLPD, both \texttt{TwinGP} and \texttt{PK} performance is comparable, and are better than the rest. Though \texttt{laGP} is the fastest at around 13 seconds with \texttt{TwinGP} being the close second at about 17 seconds, \texttt{laGP} is the worst performing model in terms of RMSE.

\subsubsection{Sarcos robotics}
\begin{figure}[ht]
\begin{center}
\includegraphics[width = \textwidth]{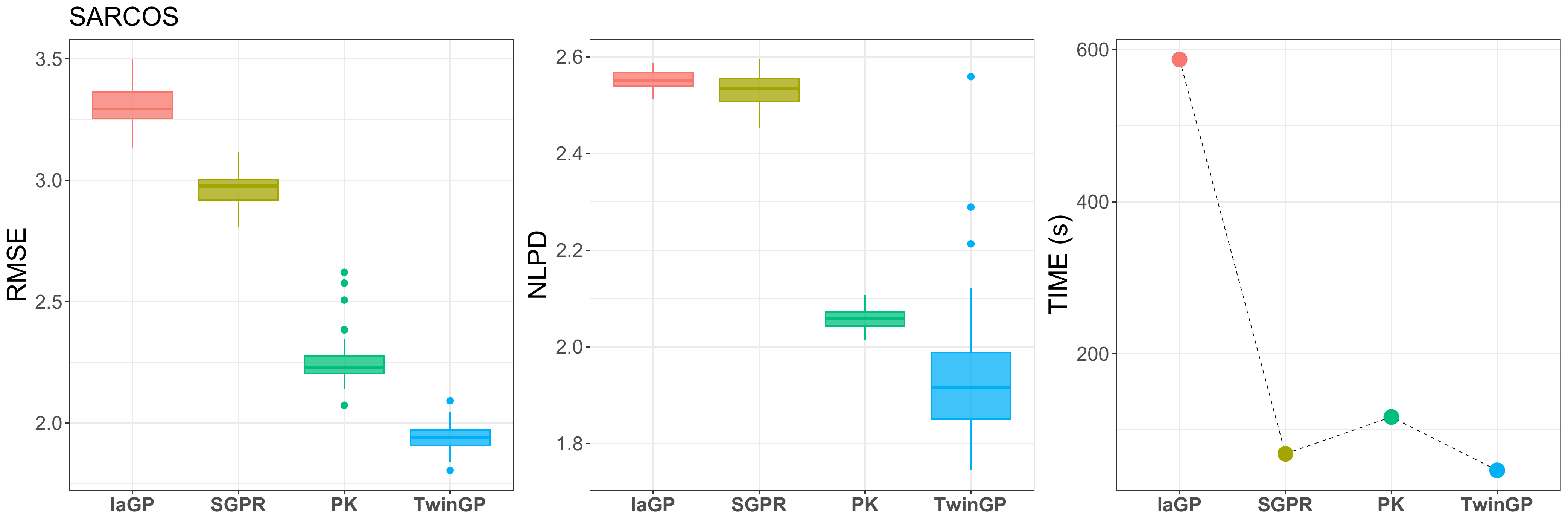} 
\caption{Distribution of RMSE and NLPD over 50 iterations of modeling the sarcos robotics dataset with \texttt{laGP}, \texttt{SGPR}, \texttt{PK}, and \texttt{TwinGP}. The corresponding average computation time per iteration is given in the right most plot.}
\label{fig:sarcos}
\end{center}
\end{figure}

The Sarcos robotics dataset\footnote{\url{http://gaussianprocess.org/gpml/data}} \citep{vijayakumar2000locally} has $d=21$ input variables representing positions, velocities, and accelerations of a seven degrees-of-freedom Sarcos anthropomorphic robot arm, and there are 7 responses  corresponding to the 7 joint torques. Similar to \cite{park2018patchwork}, we only consider the first response in modeling. The dataset is provided as training and testing sets which we combine, resulting in $48,933$ total observations, before making the 90-10 random splits. The experiment results are given in Figure \ref{fig:sarcos}. Here we see that \texttt{TwinGP} is the best performing model in terms of all the evaluation criteria. \texttt{TwinGP} computation time is only about 47 seconds.

\subsubsection{Flight delays}
\begin{figure}[ht]
\begin{center}
\includegraphics[width = \textwidth]{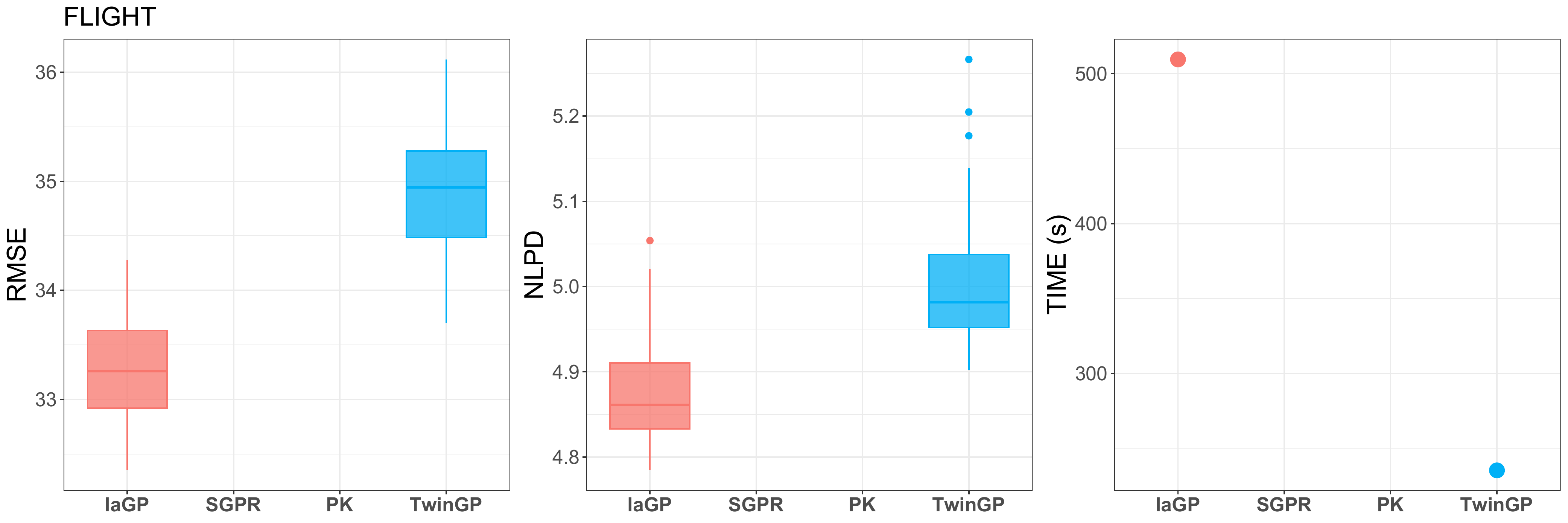} 
\caption{Distribution of RMSE and NLPD over 50 iterations of modeling the flight delays dataset with \texttt{laGP} and \texttt{TwinGP}. The corresponding average computation time per iteration is given in the right most plot.}
\label{fig:flight}
\end{center}
\end{figure}

The flight delays dataset\footnote{\url{https://community.amstat.org/jointscsg-section/dataexpo/dataexpo2009}} consists of flight arrival and departure details for all commercial flights within the USA, from October 1987 to April 2008. In keeping with previous studies involving this dataset, 800,000 observations are randomly selected for this study from a total of about 120 million observations.  Following \cite{park2018patchwork}, $d=8$ input variables are used in modeling, they are the age of the aircraft, distance that needs to be covered, airtime, departure time, arrival time, day of the week, day of the month, and month. The response is the arrival time delay. The experiment results are given in Figure \ref{fig:flight}. \texttt{SGPR} ran out of memory on our machine, and a single iteration with \texttt{PK} did not finish in one hour, hence, we have excluded both \texttt{SGPR} and \texttt{PK} from the plots. Here, \texttt{laGP} performs better than \texttt{TwinGP} in terms of RMSE and NLPD, while \texttt{TwinGP} is twice as fast.

\section{Conclusions} \label{sec:conc}
In this article, we presented a unified global-local GP approximation that addresses the drawbacks of purely global, or local approximations in the literature. With our approximation framework, referred to as \texttt{TwinGP}, GP modeling on a million data points can be performed in just a few minutes on an ordinary personal computer. The two main features of \texttt{TwinGP} are: $(i)$ the set of design points considered per testing location is a union of global points and local points, and $(ii)$ the correlation function is modeled as sum of two kernels, one each to capture the global and local trend. The set of global points are identified with \texttt{Twinning}, and are independent of the testing location, while the local points are selected as nearest neighbors to a given testing location in the training data. The training complexity of our framework is $\mathcal{O}(g^3)$ where $g$ is the number of global points used, and for $t$ testing locations, the testing complexity is $\mathcal{O}(tg^2l)$ where $l$ is the number of local points used. Our experiments show that the predictive performance with $\texttt{TwinGP}$ is on par or better than state-of-the-art GP modeling frameworks aimed at large datasets, moreover, at a fraction of their computational cost. 

\vspace{.25in}

\begin{center}
{\Large\bf Acknowledgments}
\end{center}

\noindent This research is supported by  U.S. National Science Foundation grants CMMI-1921646 and DMREF-1921873.

\bibliography{bibliography}

\end{document}